\pdfoutput=1

\documentclass[11pt]{article}

\usepackage[review]{acl}

\usepackage{times}
\usepackage{latexsym}

\usepackage[T1]{fontenc}

\usepackage[utf8]{inputenc}
\usepackage{amsmath}
\usepackage{amssymb}
\usepackage{microtype}
\usepackage{booktabs}
\usepackage{inconsolata}

\usepackage{graphicx}
\usepackage{tabularx}

\title{MCSD: An Efficient Language Model with Diverse Fusion}

\author{Hua Yang \\
  \texttt{yanghua@rockai.net} \\\And
  Duohai Li \\
  \texttt{liduohai@rockai.net} \\\And
  Shiman Li \\
  \texttt{lishiman@rockai.net} \\}

\begin{document}
\maketitle
\begin{abstract}
Transformers excel in Natural Language Processing (NLP) due to their prowess in capturing long-term dependencies but suffer from exponential resource consumption with increasing sequence lengths. To address these challenges, we propose MCSD model, an efficient language model with linear scaling and fast inference speed. MCSD model leverages diverse feature fusion, primarily through the multi-channel slope and decay (MCSD) block, to robustly represent features. This block comprises slope and decay sections that extract features across diverse temporal receptive fields, facilitating capture of both local and global information. In addition, MCSD block conducts element-wise fusion of diverse features to further enhance the delicate feature extraction capability. For inference, we formulate the inference process into a recurrent representation, slashing space complexity to $O(1)$ and time complexity to $O(N)$ respectively. Our experiments show that MCSD attains higher throughput and lower GPU memory consumption compared to Transformers, while maintaining comparable performance to larger-scale language learning models on benchmark tests. These attributes position MCSD as a promising base for edge deployment and embodied intelligence.
\end{abstract}

\section{Introduction}
Recent years have witnessed significant strides in Natural Language Processing (NLP), notably the emergence of Large Language Models (LLMs) \cite{LLM}, transforming machine-human interaction by mimicking human-like language comprehension and generation. Among them, Transformer dominates in NLP due to its powerful performance \cite{1}. Benefiting from its self-attention mechanism, Transformer has long-range dependencies that substantially improve the ability to process language. Transformer-based LLMs trained on extensive datasets sourced from the web have achieved remarkable success 
\cite{brown2020language,achiam2023gpt4,gemini2023}. However, it suffers from the disadvantage of high computational resource consumption, accompanied by $O(N^2)$ computational complexity and $O(N)$ space complexity at inference \cite{kvcache}. The computational requirements of the model scale quadratically with the length of the input sequence $N$ during inference. This limits its application to some scenarios, such as time-sensitive or resource-limited environments, typical of edge devices. 

Many attempts have been made to address the above drawbacks. Some methods simplify the query mechanism by replacing multi-head attention with multi-query attention \cite{kvcache} and group query attention \cite{ainslie2023gqa}. The memory requirement is reduced by sharing the keys $K$ and values $V$ among all heads or grouped heads, providing a flexible trade-off between computational efficiency and model representation capability. Other approaches focus on improving the computational efficiency of attention, such as AFT \cite{AFT} and RWKV \cite{2}. They obviate computing and storing expensive attention matrices by optimizing the matrix multiplication, resulting in linear computational complexity. However, these methods still use a QKV-based querying mechanism as multi-head attention does, which requires global interactions between queries and values. Recently, alternatives like Mamba \cite{mamba}, rooted in state-space model (SSM) evolution, have gained traction in the reseach community. Mamba extends input information into a higher-dimensional hidden space, demonstrating efficiency and lightness. Yet, empirical evidences suggest scaling challenges for Mamba \cite{de2024griffin}, indicating ongoing hurdles in its widespread adoption.

To this paper, we propose MCSD, an efficient language model that achieves a trade-off between performance and computational efficiency through diverse fusion. Specifically, our method facilitates both local and global feature fusion through the innovative MCSD block. In the MCSD block, it contains a slope section and a decay section, each adept at integrating long-range and short-range information, respectively. Subsequently, combining the outputs of two sections can empower the network with rich feature extraction capabilities. We use multi-channel slope and multi-channel decay to integrate historical information across varying temporal receptive fields, and leverage diverse perturbations for fine-grained element-wise integration. In addition, our approach uses multiple predefined linear operations instead of traditional dot product token interactions, thereby achieving a linear computational complexity and significantly reduced memory footprint during inference.
Our contributions in this paper are as follows:

\begin{itemize}
    
\item A new MCSD model is proposed to achieve a balance between computational consumption and representation ability via series of linear fusion, helping to address the scaling and deployment challenges.

\item To enhance the feature extraction capability, Multi-Channel Slope and Decay (MCSD) block is proposed to achieve rich feature extraction and diverse fusion, which ensure the diversity of feature extraction and fine-grained feature interaction.

\item A recurrent representation is proposed during inference stage to accelerate the inference speed. This simplified approach enjoys a computational complexity linear to the length of the sequence and has a low and stable memory complexity.

\item Experiments show that the MCSD approach outperforms Transformer on three metrics: GPU memory, latency, and throughput, showing the robust scalability of our method. The results confirm its competitive edge in delivering high-performance outcomes at low computational costs, making it a viable solution for resource-constrained edge-device deployments.
\end{itemize}

\section{Methodology}

\subsection{Architecture}\label{sec21}
\begin{figure}[]
	\centering
	\graphicspath{{fig/}}
	\includegraphics[width=0.5\textwidth]{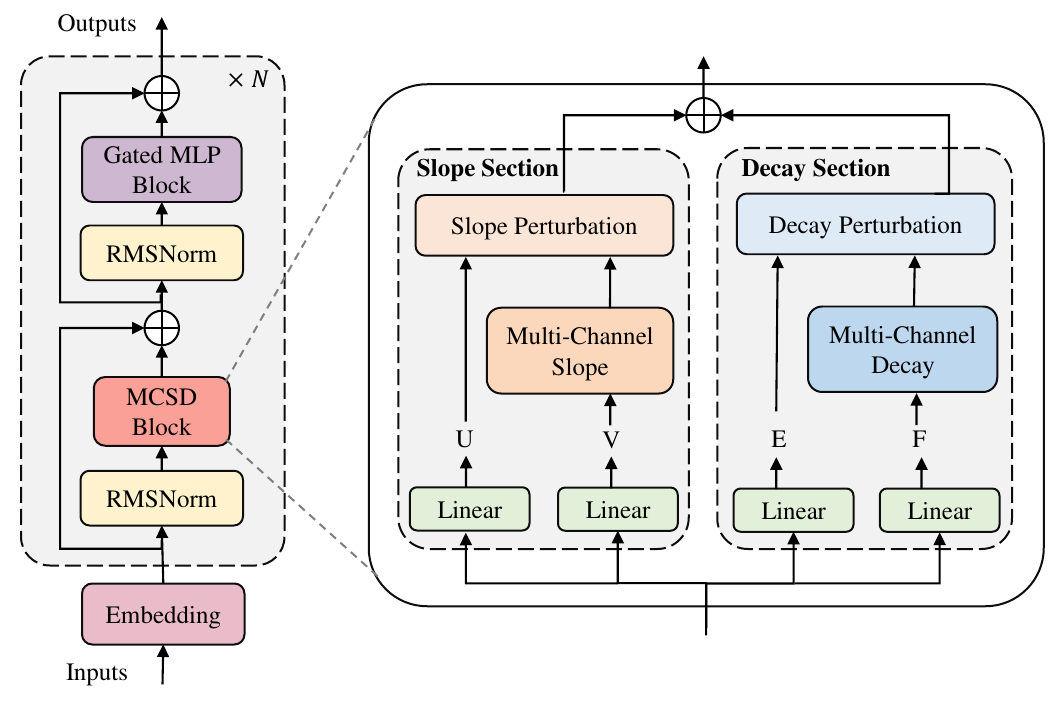}
	\caption{The architecture of our MCSD model (left). The proposed MCSD block (right) with decay and slope sections.}
	\label{model architecture}
\end{figure}

The MCSD model's architecture, depicted in Fig. \ref{model architecture} (left), is a fully-connected feed-forward network comprising $N$ stacked identical layers, as shown in Fig. \ref{model architecture} (left side). Input tokens are initially embedded to map discrete symbols into a continuous space. After embedding, the data flows into the network for subsequent feature learning and prediction of network outputs. Within each layer, two residual connections encapsulate the MCSD Block and Gated MLP Block, each preceded by RMSNorm \cite{zhang2019rmsln} for training stability enhancement. The MCSD Block, detailed in Section \ref{sec22}, constitutes the cornerstone of our innovation. The Gated MLP Block mirrors the design outlined in \cite{dauphin2017language}, utilizing GeGLU \cite{shazeer_glu_variants_2020} as the activation function in conjunction with linear layers.

\subsection{MCSD block}\label{sec22}
\begin{figure*}[t]
	\centering
	\graphicspath{{fig/}}
	\includegraphics[width=1.\textwidth]{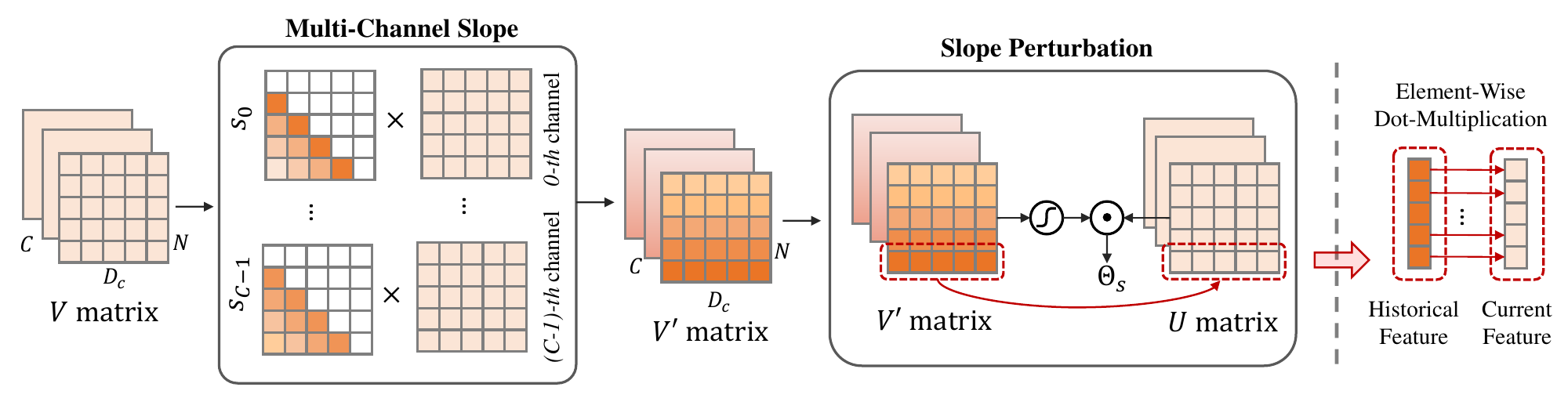}
	\caption{The slope section comprises multi-channel slope and slope perturbation, integrating past positional information via distinct slope matrices and conveying historical data to current features through element-wise multiplication, respectively. A gating mechanism filters this output, predominantly preserving current information.}
	\label{slope}
\end{figure*}
The internal structure of MCSD block is showed on the right side of Fig. \ref{model architecture}. MCSD block integrates slope and decay sections, which capture local and global feature extraction via different pre-defined weight matrix, respectively. Each section employs a dual-branch design. One branch extracts multi-channel historical features by aggregating the context from preceding tokens, another branch conducts linear projection. Two branches are followed by a perturbation operation, which facilitates the interaction between two branches. 
Specifically, given input $X$ to the MCSD block, a dimensional transformation operation is first performed $X$, ${X\in {{\mathbb{R}}^{N\times D}}}\to X\in {{\mathbb{R}}^{C\times N\times {{D}_{c}}}}$, where $N$ denotes sequence length, $D$ denotes feature dimension, $D=C\times {{D}_{c}}$, $C$ represents the number of channels, and ${D}_{c}$ represents the feature dimension of each channel. 

There are two main differences between slope and decay sections: \textit{1. Divergent Pre-defined Weights.} This determines a long or short context-awareness ability of the two in extracting historical information. \textit{2. Contrasting Perturbation Directions.} This design ensures that the slope section perturbs current features by local historical information, whereas the decay section perturbs global historical features in response to current inputs, fostering a nuanced interplay between past and current contexts.
\\
\begin{figure*}[t]
	\centering
	\graphicspath{{fig/}}
	\includegraphics[width=1.\textwidth]{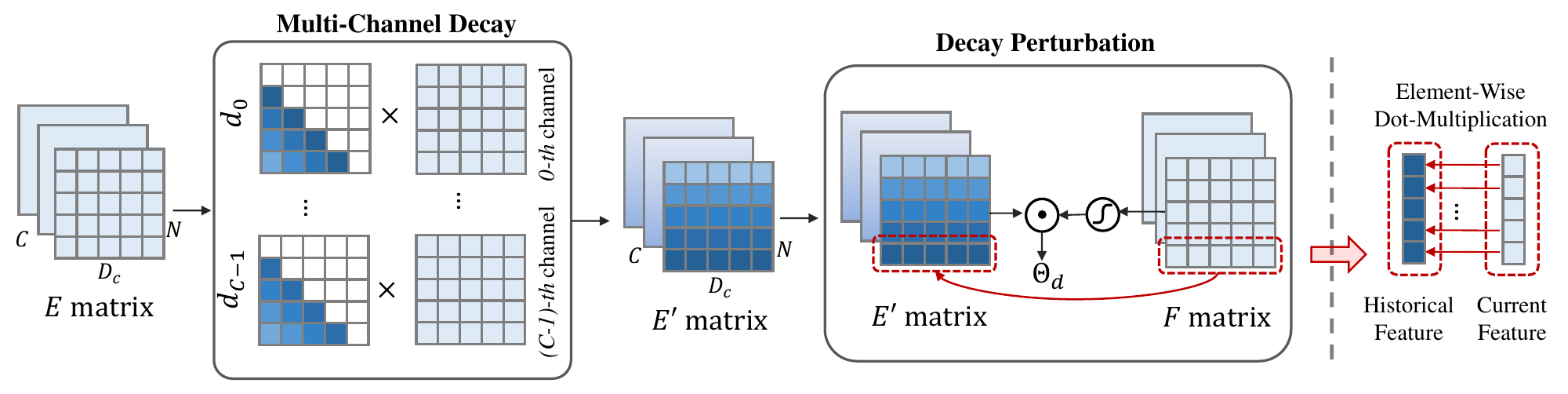}
	\caption{The decay section encompasses multi-channel decay and decay perturbation, integrating past positional data via distinct decay matrices and updating historical information through element-wise multiplication with current features. A gating mechanism selectively filters this output, primarily conserving historical information.}
	\label{decay}
\end{figure*}


\noindent
\textbf{Slope section} 
The detail of the slope section is shown in Figure \ref{slope}. Given the transformed input $X\in {{\mathbb{R}}^{C\times N\times {{D}_{c}}}}$, the slope section yields the matrix $U$,$V$ by linear projection as illustrated in Eq. (\ref{1}), where ${{W}_{U}},{{W}_{V}}\in {{\mathbb{R}}^{C\times {{D}_{C}}\times {{D}_{C}}}}$; $U,V\in {{\mathbb{R}}^{C\times N\times {{D}_{c}}}}$. 
\begin{equation}
U=X{{W}_{U}},  V=X{{W}_{V}}\label{1}
\end{equation}

In the multi-channel slope operation on matrix $V$, we set different weight matrices for different channels to realize diverse contextual feature fusion. The corresponding weight $\beta_{i}$ for each channel is defined as follows:

\begin{equation}
{{\beta }_{i}}={{\left( {{2}^{{}^{-8}/{}_{C}}} \right)}^{i+1}},i\in \left\{0,1,...C-2,C-1 \right\}\label{2}
\end{equation}
where $\beta_{i}$ is the weight applied to the $i$-th channel pair. Then, we pre-define the slope matrix $s_i$ for the $i$-th channel as shown in Eq. \ref{3}. In order to integrate the extracted historical information, we pre-define $s_i$ as a lower triangular matrix with decreasing weights in the lower triangular direction, which assigns different weights to different positions in the $N$ dimension. The upper-triangular portion of the slope matrix $s_i$ is negatively infinite. We normalize the matrix by a softmax function $\delta$ to ensure stable feature extraction under the pre-defined matrix. Since $e^{-\infty} \rightarrow 0$, the upper triangular portion of matrix $s_i$ tends to $0$ after the softmax operation, which is equivalent to mask the current and subsequent information for the historical information that does not include the current information. To prevent the denominator of the first row from approaching 0, set the first number to 1.
 
\begin{equation}
{{s}_{i}}=\delta\left(\left[ \begin{matrix}
	1  & \cdots  & \cdots  & -\infty   \\
	-{{\beta }_{i}} & -\infty  & \ddots  & \vdots   \\
	\vdots  & -{{\beta }_{i}} & -\infty  & \vdots   \\
	\left( -N+1 \right){{\beta }_{i}} & \cdots  & -{{\beta }_{i}} & -\infty   \\
\end{matrix} \right]\right)
\label{3}
\end{equation}

 The set of $C$ channel $s_i$ matrices forms the weight matrix $W_s$ for subsequent computation. Subsequently, matrix $W_s$ and matrix $V$ undergo the matrix multiplication. Here, the essence of matrix multiplication lies in the weighting and manipulation of the historical feature of input $V$, thereby imbuing the output feature with information from historical tokens. The specific operation is shown in Eq. (\ref{5}), where ${V}'\in {{\mathbb{R}}^{C\times N\times {{D}_{c}}}}$ denotes the matrix $V$ output after the multi-channel slope operation. This operation yields features ${V}'$ containing historical information.

\begin{equation}
{{W}_{s}}=\left( {{s}_{0}},{{s}_{1}},...,{{s}_{C-2}},{{s}_{C-1}} \right)\label{4}
\end{equation}

\begin{equation}
	{V}'={{W}_{s}} V\label{5}
\end{equation}

\begin{equation}
	{{\Theta}_{s}}=\sigma \left( {{V}'} \right)\odot U\label{6}
\end{equation}

Then, a slope perturbation is employed on the linear projection output $U$ and the feature output of multi-channel slope $V'$. The slope perturbation includes a gating mechanism to control the direction of perturbation and a element-wise dot-multiplication to fuse the information. The historical information $V'$ first pass the gating function and then is used to perturb feature $U$ via the dot-multiplication. The specific formula is shown in Eq. (\ref{6}). In this equation, $\sigma$ represents the SiLU function \cite{elfwing2017sigmoidweighted}, and $\Theta_{s}\in {{\mathbb{R}}^{C\times N\times {{D}_{c}}}}$ represents the output of the slope section. The slope perturbation bring the feature with historical information to the current linear input $U$ to obtain feature with adjacent information interaction.

After multi-channel slope and slope perturbation, we can get the short-range context-aware features, which incorporate the adjacent history information to the current information.
\\

\noindent
\textbf{Decay section} 
The detail of the decay section is shown in Figure \ref{decay}. Similarly, given the input data $X$, the matrics $F,E$ are obtained by the linear projection, as illustrated in Eq. (\ref{7}), where ${{W}_{F}},{{W}_{E}}\in {{\mathbb{R}}^{C\times {{D}_{C}}\times {{D}_{C}}}}$; $F,E\in {{\mathbb{R}}^{C\times N\times {{D}_{c}}}}$.

\begin{equation}
	F=X{{W}_{F}},E=X{{W}_{E}}\label{7}
\end{equation}

One branch performs a multi-channel decay operation on the matrix $E$, which is analogous to the slope section, with a different weight matrix. The multi-channel decay operation is illustrated in Eq. (\ref{8}-\ref{11}). Inspired by \cite{3}, we define different $\alpha_{i}$ values for each channel. We then incorporate these $\alpha$ values to build the triangular matrix and the triangular matrix is normalized by the activation function $\varepsilon$ by RMSNorm to obtain decay matrix $de_{i} \in {{{\mathbb{R}}}^{N\times N}}$ matrix. The formulas for $\alpha_{i}$ values of each channel and decay matrix $de_{i}$ are shown in equations (\ref{8}-\ref{9}).

\begin{equation}
	{{\alpha }_{i}}=1-{{2}^{-5-i}},i\in \left\{ 0,1,...,C-2,C-1 \right\}\label{8}
\end{equation}

\begin{equation}
	{{de}_{i}}=\varepsilon\left(\left[ \begin{matrix}
		1  & -\infty  & \cdots  & \cdots  & -\infty   \\
		\alpha _{i}^{1} & -\infty  & \ddots  & \ddots  & \vdots   \\
		\vdots  & \ddots  & -\infty  & \ddots  & \vdots   \\
		\alpha _{i}^{N-2} & \alpha _{i}^{N-3} & \ddots  & -\infty  & -\infty   \\
		\alpha _{i}^{N-1} & \alpha _{i}^{N-2} & \cdots  & \alpha _{i}^{1} & -\infty   \\
	\end{matrix} \right] \right)
	\label{9}
\end{equation}

Then ${E}'$ can be defined as shown in Eq. (\ref{10}). The $de_{i}$ matrices of the different channels are concatenated to obtain the ${W}_{d}\in {{\mathbb{R}}^{C\times N\times N}}$ matrix for the multi-channel decay operation. As shown in Eq. (\ref{11}), a multi-channel decay operation is performed on the matrix $E$ to extract and normalize the history information of the input data. Unlike the multi-channel slope, the multi-channel decay operation assigns relatively high weights to the more distant historical information, thus ensuring that the matrix ${E}'\in {{\mathbb{R}}^{C\times N\times {{D}_{c}}}}$ contains as much global historical information as possible. The $W_d$ is calculated by:

\begin{equation}
{{W}_{d}}=\left( {{de}_{0}},{{de}_{1}},...,{{de}_{C-2}},{{de}_{C-1}} \right)\label{10}
\end{equation}

\begin{equation}
	{E}'={W}_{d}E \label{11}
\end{equation}

The output of the decay section $\Theta_{d}$ is defined by Eq.(\ref{12}). A decay perturbation acts on linear projection output $F$ and multi-channel decay feature output $E'$, incorporating a gating mechanism to steer perturbation direction and an element-wise multiplication for information fusion. The gating function applied to $F$ modulates its impact on $E'$ through dot-multiplication. Here, $\sigma$ represents the sigmoid function, and $\Theta_{d}\in {{\mathbb{R}}^{C\times N\times {{D}_{c}}}}$ represents the output of the $E'$ section. The perturbation bring the input feature $F$ to feature with global information $E'$ to obtain long-dependency information.

\begin{equation}
	{{\Theta}_{d}}={E}'\odot \sigma \left( F \right)\label{12}
\end{equation}

\begin{equation}
\Theta=Concat\left( {{\Theta}_{s}},{{\Theta}_{d}} \right)\label{13}
\end{equation}

The MCSD block's output, denoted by $\Theta\in {{\mathbb{R}}^{C\times N\times {2{D}_{c}}}}$, emerges from concatenating slope and decay section outputs (Eq. \ref{13}). The combination of the two sections allows the MCSD block to extract information from closer histories without losing information from more distant histories, enhancing local and global feature extraction. 
Meanwhile, the pre-defined multi-channel matrix enriches the representation on the feature subspace.

\subsection{Fast inference for the MCSD block}\label{sec23}

The matrix multiplication in the current algorithm can benefit from the high speed parallel computation of GPUs in training. However, in inference, it increases with input length $N$ causes $O(N^2)$ algorithmic complexity and $O(N)$ space complexity consumption. This is detrimental to the efficient deployment of the model in end devices. To address this problem, we simplify the computation of model inference by a recursive form. This transformation results in a space complexity of $O(1)$ and a time complexity of $O(N)$ for the inference process, which greatly reduces the inference time. The optimized inference step in multi-channel ramp is shown in Eq. (\ref{14}-\ref{16}), while the optimized inference step in multi-channel decay is shown in Eq. (\ref{17}-\ref{19}).

\begin{equation}
	{{{V}'}_{n}}=\frac{\sum\limits_{j=1}^{n-1}{{{e}^{\left( -j \right){{\beta }_{i}}}}{{V}_{n-j}}}}{\sum\limits_{j=1}^{n-1}{{{e}^{\left( -j \right){{\beta }_{i}}}}}}\label{14}
\end{equation}

\begin{equation}
	{{{V}'}_{n+1}}=\frac{\sum\limits_{j=1}^{n}{{{e}^{\left( -j \right){{\beta }_{i}}}}{{V}_{n+1-j}}}}{\sum\limits_{j=1}^{n}{{{e}^{\left( -j \right){{\beta }_{i}}}}}} \label{15}
\end{equation}

\begin{equation}
	\begin{split}
		S_{n+1}^{slope}&={{{V}'}_{n+1}}={{e}^{\left( -1 \right){{\beta }_{i}}}}{{{{V}'}}_{n}}\frac{\sum\limits_{j=1}^{n-1}{{{e}^{\left( -j \right){{\beta }_{i}}}}}}{\sum\limits_{j=1}^{n}{{{e}^{\left( -j \right){{\beta }_{i}}}}}}+\frac{{{e}^{\left( -1 \right){{\beta }_{i}}}}{{V}_{n}}}{\sum\limits_{j=1}^{n}{{{e}^{\left( -j \right){{\beta }_{i}}}}}} \\ 
		& =(1-\frac{1}{\sum\limits_{j=0}^{n-1}{{{e}^{\left( -j \right){{\beta }_{i}}}}}})S_{n}^{slope}+\frac{1}{\sum\limits_{j=0}^{n-1}{{{e}^{\left( -j \right){{\beta }_{i}}}}}}{{V}_{n}}
	\end{split}\label{16}
\end{equation}
where $E_{j}\in {{\mathbb{R}}^{1\times {{D}_{c}}}}$ and $V_{j}\in {{\mathbb{R}}^{1\times {{D}_{c}}}}$ represent the $j_{th}$ vector representation in sequence, while $n$ denotes the number of currently inputted vector representations in sequence. $S_{n+1}^{slope}\in {{\mathbb{R}}^{1\times {{D}_{c}}}}$ and $S_{n+1}^{decay}\in {{\mathbb{R}}^{1\times {{D}_{c}}}}$ represent the next vector representation(with weight $0$ at position $n$) that is inferred under $n$ $E_{i}$ and $V_{i}$ inputs, respectively. The $i_{th}$ channel weights, $\beta_{i}$ and $\alpha_{i}$ represent the channel weights under multi-channel slope and multi-channel decay, respectively.

\begin{equation}
	{{{E}'}_{n}}=\sum\limits_{j=1}^{n-1}{\alpha _{i}^{j}}{{E}_{n-j}} \label{17}
\end{equation}

\begin{equation}
	{{{E}'}_{n+1}}=\sum\limits_{j=1}^{n}{\alpha _{i}^{j}}{{E}_{n+1-j}}\label{18}
\end{equation}

\begin{equation}
	\begin{split}
		S_{n+1}^{decay}&={{{{E}'}}_{n+1}}={{\alpha }_{i}}{{{{E}'}}_{n}}+{{\alpha }_{i}}{{E}_{n}} \\ 
		& ={{\alpha }_{i}}S_{n}^{decay}+{{\alpha }_{i}}{{E}_{n}} 
	\end{split}\label{19}
\end{equation}

In the original form (e.g., Eq. (\ref{14}-\ref{15}), Eq.(\ref{17}-\ref{18})), when reasoning about the data of the next token, it is necessary to compute all the data, which results in a significant increase in computational complexity. In optimized formulations such as Eq. (\ref{16}),  Eq. (\ref{19}), the next output token is only related to the input of the adjacent previous token, which greatly reduces the computational complexity.

\begin{equation}
	\begin{split}
		\theta_{n+1}^{i}=&U\odot \sigma \left( S_{n+1}^{slope} \right)+\\&\sigma \left( F \right)\odot RMSNorm\left( S_{n+1}^{decay} \right)
	\end{split}\label{20}
\end{equation}

\begin{equation}
	\Theta_{n+1}=\left( \theta_{n+1}^{0},\theta_{n+1}^{1},...,\theta_{n+1}^{C-1} \right)\label{21}
\end{equation}

The final output is presented in Eq. (\ref{20}) and Eq. (\ref{21}). In these equations, $\theta_{n+1}^{i}\in {{\mathbb{R}}^{1\times {{D}_{c}}}}$ represents the final output of the $(i+1)_{th}$ channel, while $\Theta_{n+1}\in {{\mathbb{R}}^{C\times 1\times {{D}_{c}}}}$ denotes the  $(i+1)_{th}$ output of the merged multi-channel sum, where $i\in \left\{ 0,1,...,C-2,C-1 \right\}$.

\section{Experiment}
\subsection{Experimental Details}
We train three models with different parameter size: 1.6B, 3B, 10B. 
The detail hyperparameters of our method is summarized in Table \ref{tab:sample}. Three models were trained, each with a different number of parameters: 1.6B, 3B and 10B. All three models were trained using a batchsize of 1152 and a learning rate of 8e-4. All experiments were performed using the AdamW optimizer \cite{scaling_law}.


\begin{table}[htbp]
	\centering
	\setlength{\abovecaptionskip}{1pt} 
	\caption{The hyperparameters of the proposed MCSD.}
	\label{tab:sample}
     \footnotesize  
	\resizebox{0.5\textwidth}{!}{%
	\begin{tabular}{ccccc}
		\toprule[1pt] 
	Size & Sequence Length & Hidde Size & Channel & Layers \\
		\midrule 
		\textbf{1.6B}     & 4096 & 2560 & 10   & 12   \\
		\textbf{3B}    & 4096  & 2560 & 10   & 20   \\
		\textbf{10B}    & 4096  & 3072 & 12   & 60   \\
		\bottomrule[1pt] 
	\end{tabular}
	}
\end{table}

\subsection{Scaling Curves}

\begin{figure}[ht]
	\centering
	\graphicspath{{fig/}}
	\includegraphics[width=0.48\textwidth]{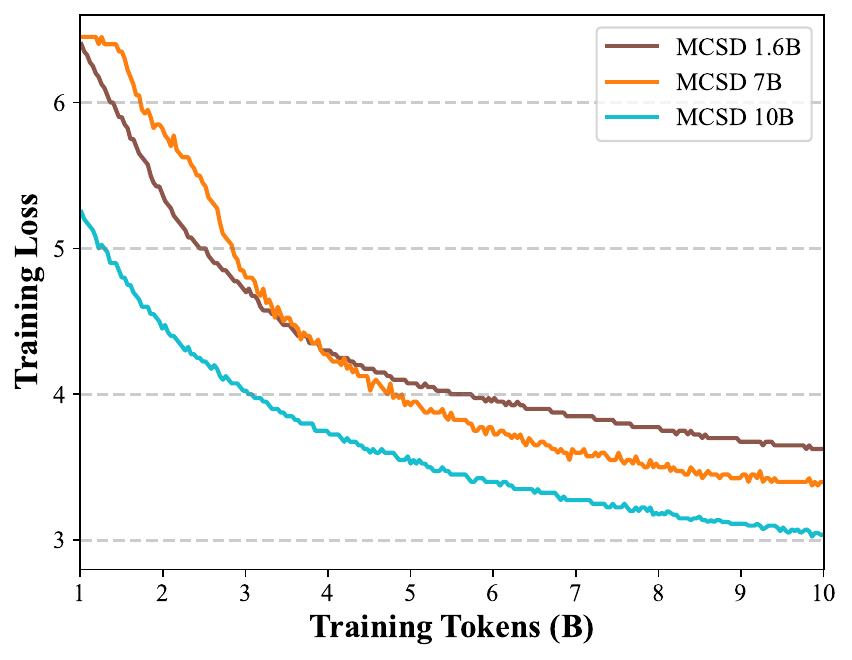}
	\caption{Scaling curves for MCSD illustrate a linear decline in loss value with growing training token volume, culminating in convergence. Larger model parameter counts correlate with diminished converged loss values.}
	\label{scaling loss}
\end{figure}

The scaling results, which demonstrate the relationship between the number of training tokens and the training loss, are presented in Figure \ref{scaling loss}. The figure illustrates three MCSD models for 1.6B, 3B and 10B, all using 4096 for the sequence length. The results present a positive relationship between training loss and model size in the case of training convergence, which is consistent with Chinchilla's scaling law \cite{adamw}. All three models exhibit convergence when the number of training tokens reaches 10B, and the models show relatively stable performance throughout the training process. The result demonstrates that the MCSD model can be trained with stability and efficiency, exhibiting good scalability.

\subsection{Inference Cost}

\begin{figure}[ht]
	\centering
	\graphicspath{{fig/}}
	\includegraphics[width=0.48\textwidth]{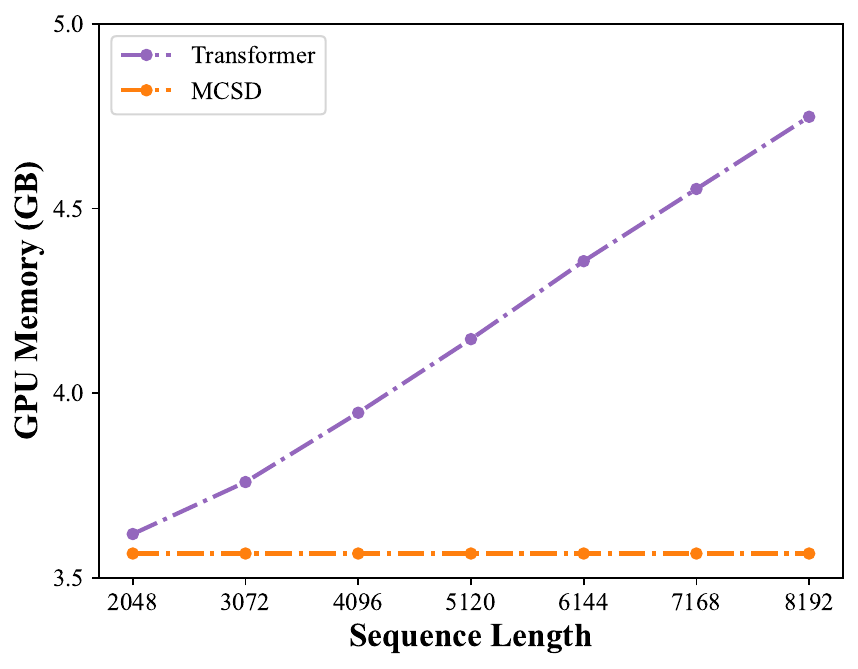 }
	\caption{GPU memory versus sequence length curves for MCSD and Transformer.}
	\label{memory}
\end{figure}

\begin{figure}[ht]
	\centering
	\graphicspath{{fig/}}
	\includegraphics[width=0.48\textwidth]{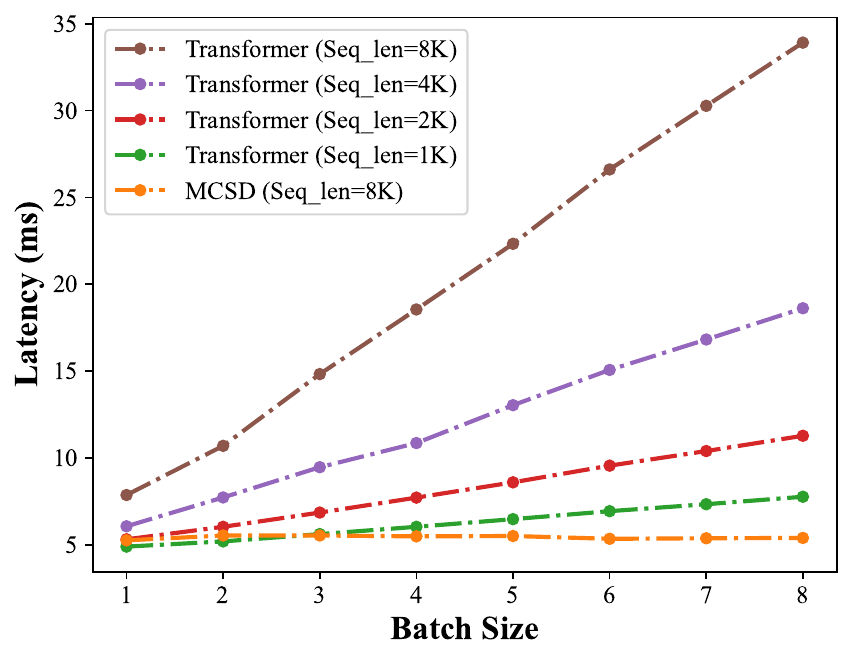}
	\caption{Latency versus batch size curves for MCSD and Transformers.}
	\label{latency}
\end{figure}

\begin{figure}[t]
	\centering
	\graphicspath{{fig/}}
	\includegraphics[width=0.48\textwidth]{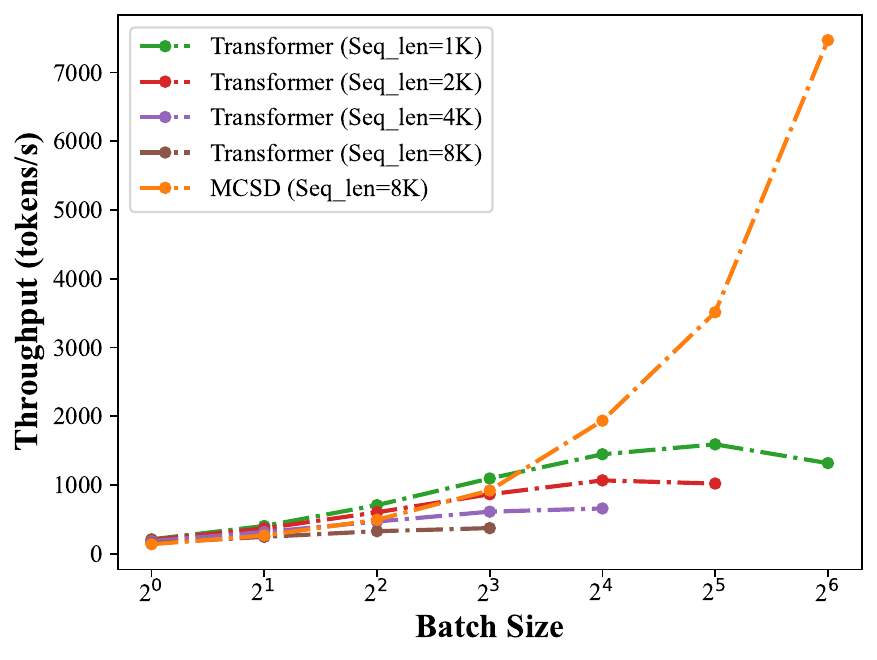}
	\caption{Throughput versus sequence length curves for MCSD and Transformers.}
	\label{throughput-seq}
\end{figure}

As illustrated in Figure (\ref{memory}-\ref{throughput-seq}), a comparison is presented between the GPU memory, latency, and throughput of the Transformer and the proposed MCSD during the inference phase. In these inference experiments, the 1.6B model was evaluated on an RTX-3090 24G GPU. The sequence length represents the number of output tokens and batch size represents the number of parallel inference processes. The prompt length is 128 tokens. 
Transformers employ the reuse of the key-value (KV) caches of previously decoded tokens. MCSD utilizes a simplified representation, as illustrated in Eq. (\ref{20}-\ref{21}). The plots demonstrate that MCSD outperforms the same parametric number of Transformer models in terms of three inference metrics.
\\

\noindent\textbf{GPU Memory}
As illustrated in Figure \ref{memory}, the memory cost of the Transformer increases linearly during inference due to the presence of the KV cache. In contrast, the memory consumption of the proposed MCSD remains almost constant even for long sequences, and MCSD has a lower memory footprint compared to the Transformer of sequence length from 2048 to 8192. This helps to achieve long sequence inference on end device with low memory footprint.
\\

\noindent\textbf{Latency}
Latency is a crucial metric in deployment, as it can significantly impact user experience. Figure \ref{latency} illustrates the latency comparison between MCSD and Transformer. The experimental results demonstrate that increasing the batch size leads to a notable increase in the delay of Transformer, which is approximately linear. Furthermore, the latency increases more rapidly when the sequence length increases. This severely constrains the applicability of Transformer to long sequence output. For a given sequence length, the delay of MCSD is considerably less than that of the Transformer, and remains relatively consistent across different batch sizes.
\\

\begin{figure}[t]
	\centering
	\graphicspath{{fig/}}
	\includegraphics[width=0.48\textwidth]{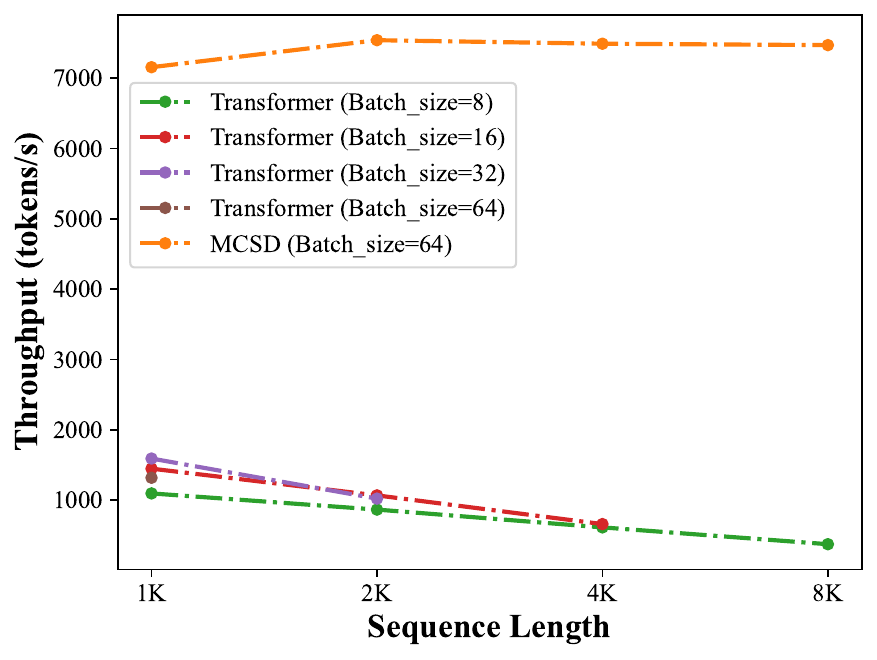}
	\caption{Throughput versus batch size curves for MCSD and Transformers.}
	\label{throughput-batch}
\end{figure}

\noindent\textbf{Throughput}
As illustrated in Figures (\ref{throughput-seq}-\ref{throughput-batch}), we examine the impact of sequence length and batch size on throughput separately. As shown in Figure \ref{throughput-seq}, we adjust the batch size for each curve, and the throughput of Transformer gradually declines with the increase in sequence length, assuming a constant batch size. In the case of a batch size of at least 16, the presence of the KV cache results in an out-of-memory issue. Conversely, the throughput of the proposed MCSD reaches a bottleneck after a slight increase in sequence length, remaining almost unchanged in the absence of out-of-memory issues. Furthermore, under the same batch size, MCSD exhibits a higher throughput performance than Transformer.

\begin{table*}[htbp]
	\centering
	\setlength{\abovecaptionskip}{1pt} 
	\caption{Comparison of the accuracy of MCSD and other models in downstream experiments with 5-shot.}
	\label{5-shot}
	\resizebox{0.85\textwidth}{!}{%
 \footnotesize
	\begin{tabular}{cccccccc}
		\toprule[1pt] 
		Model type & Model size & ARC-C   & ARC-E   & WinoGrande    & MMLU & Hellaswag &Average\\
		\midrule
		Llama3 & 8B   & 79.6 & 92.3 & 67.1 & 66.3 & 69.5 &75.0  \\
		Pythia & 2.8B & 32.9 & 64.1 & 59.7 & 26.8 & 59.3 &48.6  \\
		Mamba  & 2.8B & 36.3 & 69.7 & 63.5 & 26.2 & 66.1 &52.4  \\
		RWKV4  & 3B   & 33.1 & 67.8 & 59.6 & 25.6 & 59.6 &49.1  \\
		MCSD   & 3B   & 36.8 & 70.2 & 63.8 & 29.3 & 67.2 &53.5  \\
		\bottomrule[1pt]   
	\end{tabular}
	}
\end{table*}

The curves in Figure \ref{throughput-seq} illustrate the relationship between throughput and sequence length. The throughput of Transformer reaches a maximum and then declines with an increase in batch size. The throughput of Transformer is easily constrained and rapidly exhausts memory resources. With a fixed sequence length, the throughput of MCSD increases with the increase of batch size, and there is no out-of-memory phenomenon. Furthermore, the throughput of MCSD is higher than that of Transformer with the same sequence length. Additionally, the throughput of MCSD is several times higher than that of Transformer in the case of the same sequence length.

\subsection{Downstream Task Comparison}
A series of evaluations were conducted on the downstream tasks of baseline pre-trained models. The external baseline models we compared were Llama3-8B \cite{llama3}, Mamba-2.8B \cite{mamba}, RWKV4-3B \cite{2}, Pythia-2.8B \cite{biderman2023pythia}. Among these models, Mamba-2.8B and RWKV4-3B are the most robust small recurrent models reported in the latest literature. Pythia-2B is the most robust transformer-structured small model. Llama3 is a widely used state-of-the-art open transformer model. The proposed MCSD is trained with 3B parameters. The evaluated performance is shown in Table \ref{5-shot}. It displays the results of the 5-shot evaluation.

In Table \ref{5-shot}, a comparison was conducted between external baseline models and MCSD on MMLU, HellaSwang, ARC-E, and ARC-C, as well as WinoGrande datasets. All models were pre-trained model. The llama3-8B model is of the largest number of parameters among the models, and it outperforms the other models in all the metrics. Compared to other pre-trained model with similar model size, MCSD-3B demonstrates a high level of performance. MCSD significantly outperforms Pythia-2.8B and RWKV-3B on all metrics for all three datasets, with average metrics exceeding 4.9 and 4.4, respectively. Compared to Mamba-2.8B model, MCSD similarly outperforms it on all metrics, with an average improvement of 1.1. The above results show that among models of similar size, our proposed MCSD model has the leading performance.

\subsection{Ablation Study}

\begin{figure}[ht]
	\centering
	\graphicspath{{fig/}}
	\includegraphics[width=0.48\textwidth]{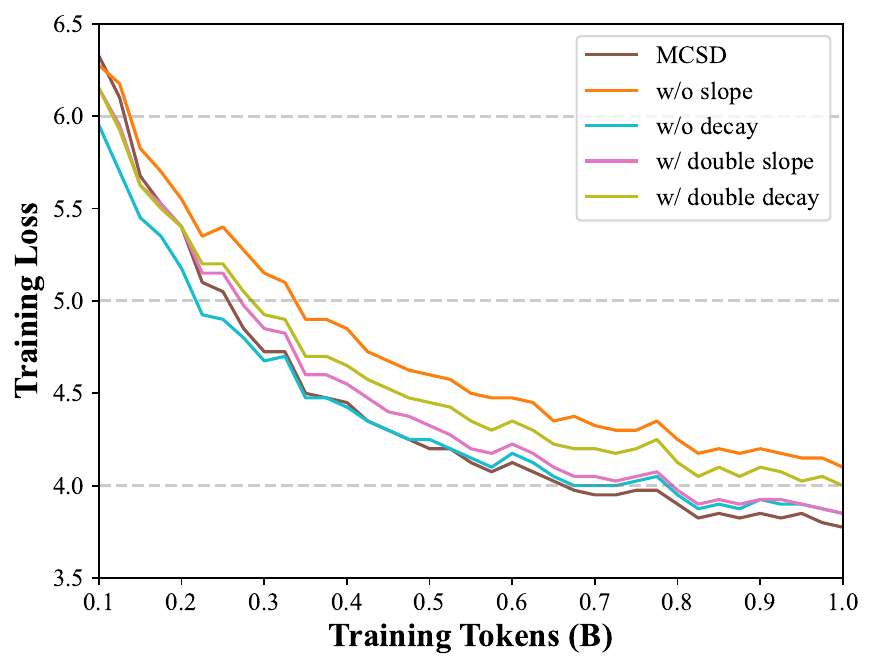}
	\caption{Ablation experiment of the proposed MCSD.}
	\label{ablation loss }
\end{figure}

Ablation experiments were conducted for both the slope section and the decay section. We established four sets of ablation experiments by either removing the decay section or the slope section, or by using two decay or two slope sections. The ablation experiment for our proposed MCSD is shown in Figure \ref{ablation loss }. The training loss is plotted as a function of the number of training tokens. The results show that deleting or replacing any of these sections increases the training loss, weakening the model effect to varying degrees. The loss of the slope section has a more pronounced impact on MCSD, while the loss of the decay section has a comparatively smaller effect. Nevertheless, this does not imply that the decay section is superfluous. Since most of the training data are short texts, it is difficult to assess the effect of the decay component on long texts.

\section{Conclusion}

This work introduces MCSD, a novel architecture that extracts the contextual information using MCSD block consisting of the slope and decay sections. This approach has been shown to enable very fast inference speed and low inference cost. During inference, MCSD exhibits better scalability and low resource consumption than Transformers with the same size. MCSD exhibits relatively superior performance compared to language modeling at all scales, showing the possibility of MCSD for end-side deployment and embodied intelligence.

\section{Limitations}
Notwithstanding advancements, limitations persist. Our method relies on publicly available corpora, which may limit its generalizability to complex professional domains such as law and medicine. Additionally, the potential presence of unmitigated toxic content within these corpora underscores toxicity mitigation as a critical area for future consideration. 
Furthermore, while showcasing reduced resource consumption, extensive deployment on edge devices awaits empirical scrutiny, demanding further assessment on hardware-limited platforms. Moreover, the current iteration of the MCSD module does not incorporate mechanisms for synchronous learning, i.e., concurrent training and inference. Enhancing our model with such capabilities could significantly broaden its applicability in real-world scenarios. Tailoring the architecture to support training-inference synchronization represents a promising avenue for future research.

\bibliography{acl_latex}




\end{document}